\documentclass[a4paper, french]{rnti}
\usepackage{algorithm, algorithmic}
\usepackage{amsfonts}
\usepackage{amssymb}
\usepackage{amsmath}
\usepackage{bbm}
\usepackage{color}
\usepackage{floatflt}
\usepackage[T1]{fontenc}
\usepackage{graphicx}
\usepackage{hyperref}
\usepackage{latexsym}
\usepackage{multirow}
\usepackage[pdftex]{pstricks}
\usepackage{tabularx}
\usepackage{times}
\usepackage{url} 
\usepackage[normalem]{ulem} 
\usepackage{verbatim}
\usepackage{wasysym}
\usepackage[normalem]{ulem}
\usepackage{subfigure}

\definecolor{greenAlex}{rgb}{0.7,1.,0.7} 
\definecolor{orangeAlex}{rgb}{1.,0.8,0.6} 
\definecolor{redAlex}{rgb}{1.,0.55,0.55}

\newcolumntype{C}[1]{>{\centering\arraybackslash}m{#1}}

\titrecourt{Construction de variables à l'aide de classifieurs comme aide à la régression}
\nomcourt{C. Troisemaine et V. Lemaire}
\titre{Construction de variables à l'aide de classifieurs \\ comme aide à la régression : une évaluation empirique}

\auteur{Colin Troisemaine\affil{1},
        Vincent Lemaire\affil{1}
}

\affiliation{
    \affil{1}Orange Labs Lannion\\ 
    colin.troisemenaine@orange.com
}

\resume{Cet article propose une méthode de création automatique de variables (pour la régression) qui viennent compléter les informations contenues dans le vecteur initial des variables explicatives. Notre méthode fonctionne comme une étape de prétraitement dans laquelle les valeurs continues de la variable a régresser sont discrétisées en un ensemble d'intervalles ce qui permet de définir des seuils de valeurs. Ensuite, des classifieurs sont entraînés pour prédire si la valeur à régresser est inférieure ou égale à chacun de ces seuils. Les sorties des classifieurs sont ensuite concaténées sous la forme d'un vecteur additionnel qui vient enrichir le vecteur initial de variables explicatives natives du problème de régression. Le système implémenté peut donc être considéré comme un outil de prétraitement générique. Nous avons testé la méthode d'enrichissement proposée avec 5 types de régresseurs et l'avons évalué dans 33 jeux de données de régression.  Nos résultats expérimentaux confirment l'intérêt de l'approche.}

\summary{This paper proposes a method for the automatic creation of variables (in the case of regression) that complement the information contained in the initial input vector. The method works as a pre-processing step in which the continuous values of the variable to be regressed are discretized into a set of intervals which are then used to define value thresholds. Then classifiers are trained to predict whether the value to be regressed is less than or equal to each of these thresholds. The different outputs of the classifiers are then concatenated in the form of an additional vector of variables that enriches the initial vector of the regression problem. The implemented system can thus be considered as a generic pre-processing tool. We tested the proposed enrichment method with 5 types of regressors and evaluated it in 33 regression datasets.  Our experimental results confirm the interest of the approach.}

\begin{document}


\section{Introduction}

Les techniques d’apprentissage peuvent se découper en deux grandes familles selon leur vocation principale : celles servant à décrire les données (méthodes descriptives) et celles permettant de prédire un phénomène (plus ou moins) observable (méthodes prédictives).
Les méthodes prédictives permettent de prévoir et d’expliquer à partir d’un ensemble de données étiquetées un ou plusieurs phénomènes (plus ou moins) observables. Dans le cas de la régression il s'agira de prévoir la valeur une variable numérique (noté $y$), par exemple le montant d'une  facture, à l'aide d'un ensemble de variables explicatives (un vecteur noté $X$).

Dans le cas de l'apprentissage automatique, on cherchera à apprendre une fonction $f$ telle que $y=f(X)$ à l'aide d'un algorithme d'apprentissage automatique et d'un ensemble d'apprentissage,  un ensemble de $N$ couples entrée-sortie  $(X_i,y_i), i={1, ..., N}$.
Lors de cette étape de modélisation, il existe souvent le besoin de créer de nouvelles variables qui décrivent mieux le problème et permettent au modèle d'atteindre de meilleures performances. C'est ce qu'on appelle le "processus d'ingénierie de création de nouvelles variables explicatives" \citep{sondhi2009feature}. Dans ce cas, on espère que les nouvelles variables (un vecteur qui sera ici noté $X'$) apporteront une information additionnelle.
L'automatisation de la génération de ces "nouvelles variables" permet d'extraire des informations plus utiles et significatives des données, dans un cadre qui peut être appliqué à n'importe quel problème. Ce qui permet à l'ingénieur en apprentissage automatique de consacrer plus de temps à des tâches plus utiles.

L’objectif de l'article est de proposer une méthode de création automatique de variables (dans le cas de la régression) qui viennent compléter les informations contenues dans le vecteur $X$ pour prédire les valeurs de la variable dépendante $y$. La méthode proposée transforme tout d'abord le problème de régression en plusieurs sous-problèmes de classification, puis intègre les résultats sous forme de variables additives ($X'$). Le vecteur 'augmenté', $X''=X \cup X'$, est ensuite mis en entrée de régresseurs usuels afin de mesurer l'apport des variables créées. L'intérêt de l'approche est présenté sous la forme d'une étude expérimentale détaillée.

\section{Proposition}
\label{sec:Proposition}


\noindent {\bf Travaux liés :} Résoudre un problème de régression en s'appuyant sur des modèles de classification est une approche qui a déjà été explorée. Ce processus a été décrit dans de nombreux articles \citep{Ahmad2012NovelEM, ahmad2018learning, TUD-KE-2010-01, jf:IJCAI-11, 8852133,TORGO1997} 
et se compose généralement de deux étapes principales : (i) la discrétisation 
de la variable cible afin de permettre l'utilisation de classifieurs sur le jeu de données; (ii) la prédiction de la régression est alors généralement réalisée en calculant la moyenne  ou la médiane des instances à l'intérieur du fragment de la sortie discrétisée que le classificateur a prédit. 
La méthode que nous proposons et présentons ci-après diffère de ces travaux car les classifieurs utilisés  par la méthode ont uniquement pour objectif d'ajouter des variables explicatives complémentaires aux variables explicatives initiales (variables natives).
Le vecteur augmenté est ensuite positionné en entrée d'un régresseur usuel. Ce régresseur prédit directement la variable cible sans nouvelle opération de transformation ou d'estimation.
Comme le montrera la section suivante, la méthode proposée est liée à une estimation conditionnelle de la fonction de densité de $y$. Il serait alors intéressant de tester d'autres méthodes de création de variables dans ce même cadre mais moins gourmandes en calcul \citep{rothfuss2019conditional,Holmes:2012,tutz2021ordinal}
dans de futurs travaux. Le principe général demeurant  le même.\\

\noindent {\bf Principe général : }\label{methode}L'idée est de tirer profit de manière opportuniste des progrès réalisés ces dernières années par les classifieurs de la littérature. Le principe de la première étape de la méthode proposée est de transformer le problème de régression en un (ou plusieurs) problème(s) de classification en \textit{discrétisant} l'espace de variation de la variable à régresser. Cette étape consiste à définir des seuils ($S$) sur l'espace de la variable à régresser.
Ces seuils seront définis à l'aide des valeurs de la variable à régresser de l'ensemble d'entraînement. Ils permettront de définir des classes  d'appartenance ({\footnotesize $C=\{C_1, ..., C_i, ..., C_S\}$}). On pourra citer comme exemple des classes définies sur des seuils d'infériorité de valeur ({\footnotesize $C_i := \mathbb{1}_{y \leq y_i}$}) ou encore des classes définies sur des appartenances à des intervalles  de valeurs ({\footnotesize $C_i := \mathbb{1}_{y \in ]y_i,y_{i+1}]}$}).
%
Une fois les classes encodées et un (ou plusieurs) classifieurs entraînés (à l'aide de l'ensemble d'entraînement) il alors possible de prédire l'appartenance des individus aux classes préalablement définies. Les prédictions du ou des classifieurs sur chacun des individus seront alors utilisées pour créer un nouvel ensemble de données "étendu", que ce soit pour l'ensemble d'entraînement ou pour l'ensemble de test.
Ainsi, en définissant $S$ seuils sur un ensemble $X$ à $d$ composantes, on obtiendra un ensemble $X''$ ayant $d+S$ composantes.  La composition du vecteur {\footnotesize $X'=\{X'_1, ..., X'_i, ..., X'_S\}$} provenant de la prédiction du ou des classifieurs sera décrite
plus loin.


Une fois le vecteur $X$ ayant été "étendu", il est possible de réaliser l'entraînement et la prédiction du modèle de régression sur ces nouvelles données. On pourra alors comparer la performance du régresseur lorsqu'on l'entraîne sur l'ensemble de données original (muni uniquement de $X$) à sa performance lorsqu'on utilise l'ensemble étendu ($X''=X \cup X'$). L'hypothèse étant que le modèle de régression aura une meilleure intuition de la position générale des individus dans l'espace de la variable à régresser et présentera de meilleurs résultats.
Le processus suivi par la méthode proposée peut être résumé en trois étapes : (i) la première  consiste à transformer le problème de régression en problème de classification. Pour ce faire, la variable cible $y$ est d'abord discrétisée, puis les classes sont définies à l'aide des seuils ainsi répartis. (ii) Dans une deuxième étape, les classifieurs sont entraînés en utilisant les variables descriptives initiales de $X$ et les nouvelles classes dérivées de $y$. Puis la prédiction des classifieurs utilisant le vecteur $X$ initial est utilisée pour extraire de nouvelles caractéristiques, c'est-à-dire $X'$. (iii) Enfin, le modèle de régression peut être entraîné à l'aide du vecteur $X''=X\cup X'$.
La méthode proposée repose donc sur un mécanisme de discrétisation et d'encodage de classes. Ces deux processus peuvent être mis en place de différentes façons. Nous proposons un exemple d'implémentation ci-après.\\ 

\noindent {\bf Implémentation : }L'un des aspects les plus importants de la méthode proposée est l'association de classes aux instances. Ces classes seront utilisées lors de l'apprentissage des classifieurs. Ce processus repose sur deux étapes : la définition de seuils et l'encodage des classes.
Pour la définition du {\it placement des seuils}, ce qui revient à une discrétisation non-supervisée de la variable à régresser, il existe dans la littérature de nombreuses possibilités telles que "EqualWidth", "EqualFreq", etc. Pour {\it l'encodage des classes}, il est possible de poser le problème soit comme un problème de classification à $S$ classes (où $S$ désigne le nombre de seuils), soit comme $S$ problèmes de classification binaire, soit comme un problème de classification multi-labels.
Les choix de la méthode de discrétisation et du nombre de seuils sont liés et ce choix représente un compromis entre (i) l'efficacité en classification et (ii) l'apport d'information pour le problème initial de régression. Dans le cas d'un unique classifieur qui prédirait l'intervalle d'appartenance des individus, il semble évident que plus S sera grand, plus l'information apportée au régresseur sera précise, mais plus le problème deviendra difficile à apprendre. En revanche, si on pose le problème comme $S$ problèmes de classification binaire, chaque classifieur aura un problème de même difficulté à résoudre, indépendamment de la valeur de $S$. Des tests préliminaires effectués sur l'ensemble des jeux de données décrits Section \ref{datasets} ont permis de vérifier ce comportement.

Pour la {\bf méthode de discrétisation} - La méthode "Equal Frequency" a été choisie car contrairement à "Equal Width", elle ne risque pas de créer d'intervalles qui ne contien\-nent aucun individu. Elle permet également de s'assurer de ne pas poser problème de classification où la classe minoritaire (seuil extrême à gauche) représente moins de $\frac{1}{S+1}$ pourcent des individus.

Pour la nature du {\bf problème de classification} - Lors des tests préliminaires, le fait de poser le problème en $S$ problèmes de classification binaire a donné de meilleures performances, tant en apprentissage qu'en déploiement. Les résultats ont aussi été plus robustes (meilleure généralisation), ce qui est un point important si l'on veut que les variables ajoutées soient bénéfiques au régresseur. Les classifieurs utilisés dans le reste de cet article sont les Forêts Aléatoires \citep{RandomForestsBreiman} de Scikit-Learn 
avec 100 arbres (mais tout autre classifieur performant et robuste pourrait être utilisé) avec leurs paramètres par défaut.

Pour le {\bf nombre de seuils} - La première intuition (qui a été confirmée lors de l'expérimentation) est que plus le nombre de seuils défini est grand, plus le gain de performance est important. En effet, plus on définit de seuils, plus la discrétisation est fine et plus la prédiction des classifieurs s'approchera de la valeur réelle de régression. Lors la phase expérimentale de cet article, nous illustrerons la performance de notre méthode en fonction du nombre de seuils défini. Le  nombre  de  seuils  défini  sur  l’espace  de  la  variable  à  régresser  est  donc  un  hyper-paramètre de notre méthode.

Pour la définition des {\bf classes associées aux seuils} - Le choix retenu s'est porté  sur des classes liées à des seuils d'infériorité de valeur ($C_i := \mathbb{1}_{y \leq y_i}$). 

Pour les {\bf variables extraites} des classifieurs - Chaque classifieur sera entraîné à prédire si les données que l'on lui fournit sont inférieures ou supérieures au seuil auquel une classe lui est associée. La méthode se voulant générique, il a été décidé d'extraire les probabilités conditionnelles prédites par chaque classifieur. En effet, cette information est accessible pour une large majorité des classifieurs de la littérature. Les probabilités conditionnelles\footnote{Note: les classifieurs retenus étant des forêt aléatoires, la probabilité conditionnelle correspondra à la proportion d’arbres ayant voté pour la classe concernée.} de la classe 1 (c'est à dire si $y \leq y_i$) prédites par chaque classifieur composeront donc le vecteur {\footnotesize $X'=\{X'_1, ..., X'_i, ..., X'_S\}=\{P(C_1=1|X), ..., P(C_i=1|X), ..., P(C_S=1|X)\}$}; $X'$ qui sera donc ajouté au vecteur de données initial ($X$). 






\section{Protocole experimental}

\label{datasets}

{\bf $\bullet$ Jeux de données: }Pour réaliser l'analyse de la méthode proposée, nous avons sélectionné une large collection de jeux de données de régression provenant du UCI Repository \citep{Dua:2019} et de Kaggle. 
33 jeux de données ont été utilisés, parmi lesquels 23 sont constitués de plus de 10 000 individus. Les 10 jeux de données restants  allant de 1 030 à 9 568 individus. Cette sélection a été
influencée par \citep{FERNANDEZDELGADO2019}. Les jeux de données choisis sont d'après la catégorisation de Fernandez-Delgado et al. "{\it grands et difficiles}". 
%

{\bf $\bullet$ Prétraitements:} A l'identique de \citep{FERNANDEZDELGADO2019}, deux prétraitements ont été réalisés avant d'entrer dans le processus décrit 
Section \ref{methode} : (i) le recodage des variables catégorielles à l'aide d'un codage disjonctif complet; (ii) la suppression des variables de dates, des variables constantes, des identifiants d'individus, des variables colinéaires et des autres variables potentiellement à 'régresser'. Trois autres prétraitements ont été réalisés sur chaque "fold" d'apprentissage. Ces derniers sont dans l'ordre de réalisation : (i)  normalisation des variables numériques (centrage - réduction); (ii) normalisation de la variable à régresser (centrage - réduction)  puis transformation à l'aide de Box-Cox \citep{10.1214/20-STS778}; (iii) création des seuils pour la définition des classes tel que décrit précédemment. Pour chacun de ces trois prétraitements, les statistiques qui leurs sont associées ont été calculées sur l'ensemble d'entraînement puis appliquées sur les ensembles d'entraînement et de test. Dans la présentation ci-après des résultats, les résultats en RMSE 
seront donnés sans effectuer la transformation inverse de la fonction Box-Cox. Enfin les examples à valeurs manquantes ont été retirés du jeu de données initial. 


{\bf $\bullet$ Découpage Train-Test et optimisation des modèles:}
Chacun des jeux de données a été découpé en un "10-fold cross validation" permettant d'avoir 10 jeux d'entraînement et de test. Pour les trois modèles nécessitant une optimisation de leurs paramètres (DT, FA et XGB) : 30\% de l'ensemble d'entraînement a été réservé pour optimiser les paramètres du modèle dans un processus de "grid-search", permettant ainsi souvent d'éviter un sur-apprentissage. 
Puis, muni des "bons" paramètres d'apprentissage, le modèle a été entraîné à l'aide des 70\% restant de l'ensemble d'entraînement. Enfin, une fois le modèle appris pour ce fold, ses performances sont mesurées. 
Ce processus rigoureux a été réalisé pour tous les jeux de données, tous les folds et toutes les valeurs de $S$, résultant ainsi en l'apprentissage de milliers de modèles, mais permettant un test rigoureux de la méthode proposée.

{\bf $\bullet$ Jeux de données :}  33 jeux de données provenant de UCI ou de Kaggle ont été sélectionés . Leur nombre d'individus est compris entre 1,030 et 955,167 et leur nombre d'attributs va de 3 à 384. 

{\bf $\bullet$ Métrique utilisée pour les résultats :} 
Dans la suite des résultats, la RMSE a été choisie et est définie telle que :
{\footnotesize $RMSE = \sqrt{\frac{1}{n}\sum_{i=1}^{n}(y_i-\hat{y}_i)^2}$}, où $n$ est le nombre d'individus, $y_i$ la valeur désirée et $\hat{y}_i$ la sortie d'un regrésseur.

{\bf $\bullet$ Régresseurs testés:} Lors des expérimentations, cinq régresseurs utilisant des cadres différents ont été utilisés.
Étant largement connus par la communauté de l'apprentissage automatique, nous ne les décrirons pas. Notre choix s'est porté sur la régression linéaire (LR), l'arbre de régression (DT), la forêt aléatoire (RF), XGBoost (XGB) et enfin le Na\"ive Bayes Pondéré (SNB). On notera cependant que le Na\"ive Bayes Pondéré qui a été utilisé ici a été produit par le logiciel Khiops ({\url{www.khiops.com}}). Nous avons demandé et obtenu gratuitement une licence provisoire du logiciel, ainsi que sa version pyKhiops utilisable dans Scikit-Learn. Nous avons utilisé la version 1.4.1 de XGBoost et la version 0.24.2 de Scikit-Learn pour les trois méthodes de régression restantes.

\section{Résultats}

\begin{figure}[!ht]
    \centering
    \includegraphics[width=0.85\linewidth]{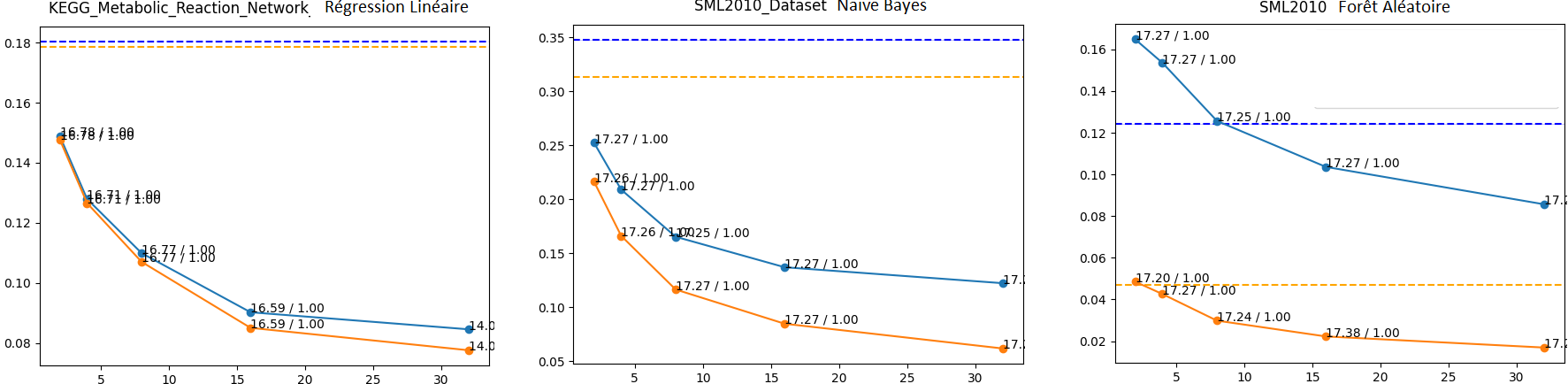}
    \caption{$S$ (axe horizontal) versus RMSE (axe vertical). Les lignes 
    en pointillés représentent la performance initiale du régresseur
    en bleu pour l'ensemble de test et en orange pour l'ensemble d'entraînement. Les courbes en trait plein représentent la performance du régresseur avec la méthode proposée, 
    en reprenant le même code couleur.
    }
    \label{fig:example}
\end{figure}

{\bf $\bullet$  Résultats\footnote{Le lecteur pourra trouver la description complète des jeux de données, le code permettant de reproduire l'ensemble des résultats et une version étendue de cet article sur \url{https://github.com/ColinTr/ClassificationForRegression}.} illustratifs - } La Figure \ref{fig:example} illustre le comportement de la méthode proposée : A gauche pour la régression linéaire et le jeu de données "KEGG Metabolic Reaction", au milieu et à droite pour le jeu de données "SML 2010" respectivement pour le régresseur naïf de Bayes et la forêt aléatoire. Cette figure est  représentative des résultats obtenus, avec une décroissance plus ou moins marquée de la RMSE versus la valeur de $S$ et le régresseur considéré.  On observe que le nombre de seuils $S$  influe sur le processus d'apprentissage. Cependant il a été constaté que le gain de performance est proportionnel au nombre de seuils définis : une décroissance de la RMSE versus la valeur de $S$ est apparente; plus marquée en début de courbe et plafonne ensuite pour un gain moins marqué au-delà de 16 seuils.

\begin{table}[!ht]
    \centering
    \fontsize{7}{7}\selectfont
    \setlength{\tabcolsep}{3pt}
    \begin{tabular}{|l|c|c|c|c|c|c|c|c|c|c|}
        \hline
        & \multicolumn{2}{c|}{\textbf{Linear Regr.}} & \multicolumn{2}{c|}{\textbf{Decision Tree}} & \multicolumn{2}{c|}{\textbf{Random Forest}} & \multicolumn{2}{c|}{\textbf{XGBoost}} & \multicolumn{2}{c|}{\textbf{SNB}} \\
        \hline
        \textbf{Dataset name} & \textbf{Natif} & \textbf{Aug} & \textbf{Natif} & \textbf{Aug} & \textbf{Natif} & \textbf{Aug} & \textbf{Natif} & \textbf{Aug} &\textbf{Natif} & \textbf{Aug} \\
        \hline
        3Droad               & 0,9867 & \textbf{0,0930} & 0,1213 & \textbf{0,0831} & 0,0908 & \textbf{0,0792} & 0,0935 & \textbf{0,0781} & 0,5580 & \textbf{0,0827} \\
        air-quality-CO       & 0,3269 & \textbf{0,2672} & 0,3255 & \textbf{0,2727} & 0,2667 &          0,2642 & 0,2682 &          0,2633 & 0,3428 & \textbf{0,2768} \\
        airfoil              & 0,7176 & \textbf{0,2351} & 0,4373 & \textbf{0,3194} & 0,2987 &          0,2865 & 0,2814 &          0,2752 & 0,7709 & \textbf{0,2963} \\
        appliances-energy    & 0,8260 & \textbf{0,4865} & 0,7431 & \textbf{0,5267} & 0,5456 & \textbf{0,5164} & 0,5790 & \textbf{0,5203} & 0,8354 & \textbf{0,5206} \\
        beijing-pm25         & 0,7833 & \textbf{0,4180} & 0,6013 & \textbf{0,3948} & 0,4168 & \textbf{0,3874} & 0,4149 & \textbf{0,3885} & 0,8082 & \textbf{0,3931} \\
        temp-forecast-bias   & 0,4749 & \textbf{0,2847} & 0,4668 & \textbf{0,2848} & 0,3237 & \textbf{0,2771} & 0,3085 & \textbf{0,2773} & 0,4782 & \textbf{0,2907} \\
        bike-hour            & 0,7163 & \textbf{0,2547} & 0,3294 & \textbf{0,2564} & \textbf{0,2415} & 0,2499 & \textbf{0,2208} & 0,2500 & 0,4858 & \textbf{0,2534} \\
        blog-feedback        & 0,8307 & \textbf{0,6434} & 0,6741 &          0,6694 & \textbf{0,6481} & 0,6618 & \textbf{0,6444} & 0,6652 & 0,7013 & \textbf{0,6601} \\
        buzz-twitter         & 0,7248 & \textbf{0,2183} & 0,2299 & \textbf{0,2265} & \textbf{0,2170} & 0,2200 & \textbf{0,2162} & 0,2205 & 0,2238 & \textbf{0,2191} \\
        combined-cycle       & 0,2702 & \textbf{0,1967} & 0,2647 & \textbf{0,2014} & 0,2079 &          0,1952 & 0,2041 &          0,1950 & 0,2570 & \textbf{0,1995} \\
        com-crime            & 0,6312 &          0,5485 & 0,6418 & \textbf{0,5552} & 0,5525 &          0,5505 & 0,5665 &          0,5503 & 0,5612 & 0,5523 \\
        com-crime-unnorm     & 0,6751 &          0,6031 & 0,7036 & \textbf{0,6251} & 0,6163 &          0,6157 & 0,6186 &          0,6124 & 0,6453 & \textbf{0,6340} \\
        compress-stren       & 0,6331 & \textbf{0,2732} & 0,4519 & \textbf{0,3173} & 0,3137 &          0,2896 & 0,2790 &          0,2955 & 0,5962 & \textbf{0,2959} \\
        cond-turbine         & 0,3002 & \textbf{0,1096} & 0,1764 & \textbf{0,1094} & 0,1117 &          0,1064 & 0,1062 &          0,1094 & 0,4481 & \textbf{0,1096} \\
        cuff-less            & 0,7996 & \textbf{0,5791} & \textbf{0,5165} & 0,6255 & \textbf{0,5025} & 0,6105 & \textbf{0,5012} & 0,6225 & \textbf{0,5450} & 0,5922 \\
        electrical-grid-stab & 0,5961 & \textbf{0,3156} & 0,5545 & \textbf{0,3245} & 0,3352 & \textbf{0,3068} & \textbf{0,2634} & 0,3092 & 0,5951 & \textbf{0,3157} \\
        facebook-comment     & 0,6828 & \textbf{0,4242} & 0,4655 & \textbf{0,4429} & \textbf{0,4162} & 0,4384 & \textbf{0,4095} & 0,4397 & 0,5014 & \textbf{0,4347} \\
        geo-lat              & 0,8752 & \textbf{0,8066} & 0,9594 & \textbf{0,8538} & 0,8321 &          0,8287 & 0,8540 &          0,8353 & 0,8668 & 0,8408 \\
        greenhouse-net       & 0,6492 & \textbf{0,5418} & \textbf{0,5397} & 0,5686 & \textbf{0,5158} & 0,5638 & \textbf{0,5137} & 0,5618 & 0,6016 & \textbf{0,5599} \\
        KEGG-reaction        & 0,1803 & \textbf{0,0845} & 0,0919 & \textbf{0,0861} & 0,0837 &          0,0835 & 0,0835 &          0,0840 & 0,1172 & \textbf{0,0855} \\
        KEGG-relation        & 0,6342 & \textbf{0,1750} & 0,2454 & \textbf{0,1748} & 0,1782 & \textbf{0,1700} & 0,1809 & \textbf{0,1715} & 0,4314 & \textbf{0,1738} \\
        online-news          & 1,1822 &          1,0677 & 0,9516 & 0,9622          & \textbf{0,9177} & 0,9521 & \textbf{0,9124} & 0,9521 & \textbf{0,9228} & 0,9942 \\
        video-transcode      & 0,3975 & \textbf{0,0561} & 0,0846 & \textbf{0,0652} & 0,0604 &          0,0597 & 0,0590 & \textbf{0,0568} & 0,3704 & \textbf{0,0627} \\
        pm25-chengdu-us-post & 0,8022 & \textbf{0,4283} & 0,5578 & \textbf{0,4164} & 0,4111 &          0,4084 & 0,4125 &          0,4079 & 0,7716 & \textbf{0,4131} \\
        park-total-UPDRS     & 0,9472 & \textbf{0,7816} & 0,9416 & \textbf{0,8057} & 0,8020 &          0,7893 & 0,8196 &          0,7860 & 0,9428 & \textbf{0,8117} \\
        physico-protein      & 0,8453 & \textbf{0,5179} & 0,7676 & \textbf{0,5338} & 0,5528 & \textbf{0,5218} & 0,5744 & \textbf{0,5249} & 0,8494 & \textbf{0,5330} \\
        production-quality   & 0,4872 & \textbf{0,2807} & 0,3886 & \textbf{0,2830} & 0,2836 &          0,2779 & 0,2794 &          0,2781 & 0,3790 & \textbf{0,2832} \\
        CT-slices            & 1,9533 &          0,0504 & 0,1332 & \textbf{0,0586} & 0,0544 & \textbf{0,0418} & 0,0693 & \textbf{0,0355} & 0,1104 & \textbf{0,0376} \\
        gpu-kernel-perf      & 0,6347 & \textbf{0,0696} & \textbf{0,0300} & 0,0557 & \textbf{0,0246} & 0,0501 & \textbf{0,0228} & 0,0500 & 0,5927 & \textbf{0,0620} \\
        SML2010              & 0,2536 & \textbf{0,1157} & 0,2179 & \textbf{0,1036} & 0,1241 & \textbf{0,0857} & 0,1058 & \textbf{0,0920} & 0,3473 & \textbf{0,1221} \\
        seoul-bike-sharing   & 0,6990 & \textbf{0,4839} & 0,5801 & \textbf{0,5217} & 0,5101 &          0,5149 & 0,5171 &          0,5139 & 0,5891 & \textbf{0,5169} \\
        uber-location-price  & 0,9998 & \textbf{0,4589} & 0,6153 & \textbf{0,4789} & 0,4635 &          0,4699 & 0,4705 &          0,4688 & 0,8419 & \textbf{0,4879} \\
        year-prediction      & 0,8569 & \textbf{0,8079} & 0,8783 & \textbf{0,8137} & 0,8059 &          0,8062 & \textbf{0,7929} & 0,8131 & 0,8822 & \textbf{0,8138} \\
        \hline
        \textbf{Moyenne } & 0,7083 & 0,3842	& 0,4754 & 0,3945 & 0,3856 & 0,3842 & 0,3831 & 0,3850 & 0,5749 & 0,3917 \\

        \hline
        \multicolumn{1}{|r|}{\textbf{Déf / Egal / Vict }} & \multicolumn{2}{|c|}{ 0 / 4 / 29 }& \multicolumn{2}{|c|}{ 3 / 2 / 28 } & \multicolumn{2}{|c|}{ 8 / 16 / 9 } & \multicolumn{2}{|c|}{ 10 / 14 / 9} & \multicolumn{2}{|c|}{ 2 / 2 / 29 } \\
        \hline
        
    \end{tabular}
    \caption{Résultats pour chaque jeu de données en test (résultats moyen sur les 10 "fold" de test (voir Section \ref{datasets})) pour $S=32$.}
    \label{table:RMSEComparison}
\end{table}

{\bf $\bullet$ Tables de résultats sur l'ensemble des jeux de données - } Les résultats obtenus sont présentés en détail dans la Table \ref{table:RMSEComparison} pour un nombre de seuils égal à 32 ($S=32$).
Pour chaque jeu de données, cette table donne les résultats en test (résultats moyen sur les 10 "fold" de test (voir Section \ref{datasets})) pour chacun des cinq régresseurs pour lesquels l'ajout du vecteur $X'$ a été testé.
Dans cette table, une valeur en gras indique une différence significative entre les résultats avec le vecteur $X$ "Natif" et le vecteur $X'$ "Aug" (Augmenté) selon test de Student apparié (p-value à 5\%). La dernière ligne du tableau indique le nombre de défaites, égalités et victoires de "Aug" vis-à-vis de "Natif". On s'aperçoit que l'ajout du vecteur $X'$  profite essentiellement à 3 des régresseurs : La régression linéaire, l'arbre de régression et le régresseur naïf de Bayes. Pour la forêt aléatoire et XGboost, les gains et/ou pertes sont assez limités. 
L'avant dernière ligne du tableau présente la RMSE moyenne obtenue sur l'ensemble des jeux de données (purement à titre indicatif) confirmant  l'intérêt de la méthode proposée pour trois des cinq régresseurs. 
Cette moyenne doit être regardée avec précaution. Comme chaque ensemble de données est un problème d'une difficulté unique, l'échelle des erreurs diffère entre chaque ensemble de données. Ainsi, si la nouvelle RMSE {\it moyenne} en test est plus élevée, cela ne signifie pas nécessairement que les régresseurs ont de moins bonnes performances en moyenne. C'est pourquoi nous présentons ci-après dans la section suivante une autre vue sur les résultats afin de poursuivre leur analyse.

\begin{figure}[!ht]
	\begin{center}
		\includegraphics[width=0.85\linewidth]{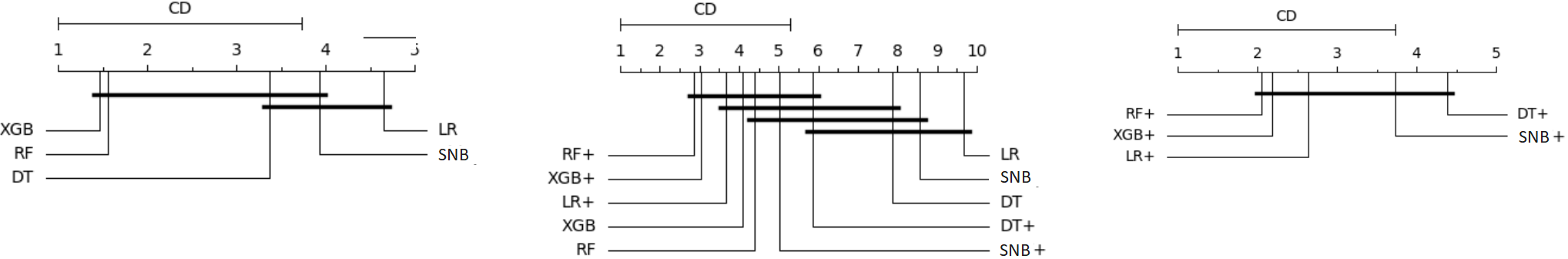}
		\caption{
		À gauche les régresseurs sans la méthode proposée, au centre les régresseurs avec et sans la méthode proposée et à droite les régresseurs uniquement avec la méthode proposée
		}
		\label{fig:cd}
	\end{center}
\end{figure}

{\bf $\bullet$ Diagramme Critique - } Nous présentons Figure \ref{fig:cd} des diagrammes critiques comparant entre eux les résultats obtenus par chaque régresseur par le biais du test post-hoc de Nemenyi, effectué après un test des rangs signés de Friedman des RMSE. 

On peut observer sur le diagramme de gauche de la figure \ref{fig:cd} que le régresseur XGBoost  présente les meilleures performances en obtenant le meilleur rang moyen de RMSE, suivi de près par la forêt aléatoire (RF). Ces deux régresseurs sont largement plus performants que les trois autres régresseurs, la régression linéaire est le modèle le moins performant. Elle est suivie de près par le régresseur naïf de Bayes et l'arbre de régression.
%
Sur la figure \ref{fig:cd} au centre : le premier point
que l'on peut observer est que statistiquement, tous les régresseurs munis de $X'$ ont un meilleur rang moyen que leur 
muni uniquement de $X$, ceci est un résultat encourageant : aucun régresseur n'a été négativement affecté par la méthode proposée, puisque tous les régresseurs qui ont utilisé cette méthode sont mieux classés que leurs versions de base. Différents groupes sont à nouveau présents et le lecteur pourra comparer le changement de classement des différents régresseurs en comparant la figure de gauche à celle du milieu.
Enfin,  le diagramme de droite de la figure \ref{fig:cd} montre que, lorsque la méthode proposée est utilisée, les régresseurs ne sont plus différenciables (un seul groupe comparé à la figure de gauche) selon le test de Nemenyi, même s'ils ont des rangs moyens différents.

\section{Conclusion}
\label{sec:Conclusion}

Cet article a proposé  une méthode de création automatique de variables (dans le cas de la régression) qui viennent compléter les informations contenues dans le vecteur initial d’entrée, les variables explicatives. 
Les résultats sont encourageants, même s'ils sont surtout bénéfiques à trois des cinq régresseur pour lesquels la méthode a été testée.
Une première amélioration serait d'extraire des classifieurs un vecteur plus informatif, par exemple les identifiants des feuilles des arbres.
Une deuxième amélioration pourrait être de concevoir une architecture neuronale regroupant l'ensemble des étapes de la méthode proposée.
La méthode proposée ouvre aussi certaines perspectives. Le lecteur attentif aura
remarqué que tel que défini dans l'implémentation proposée, le vecteur {\footnotesize $X'$} est en fait une estimation de la densité cumulée conditionnelle de $y$ connaissant $X$. Il serait donc possible via un calcul d'espérance de se passer totalement des régresseurs (si $S$ était suffisamment élevé) pour avoir une assez bonne estimation de cette densité cumulée. Ce dernier point fera probablement l'objet de travaux futurs.

\bibliographystyle{rnti}
\bibliography{biblio}

\section*{Annexe}

\subsection*{A-1 : Illustration de la définition des seuils}

\begin{figure}[h]
	\begin{center}
		\includegraphics[width=0.45\textwidth]{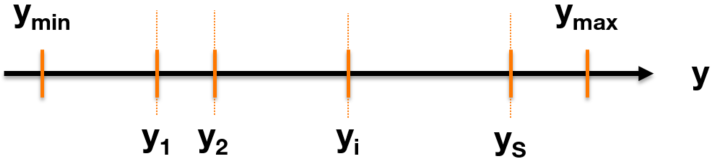}
		\caption{Exemple de seuils}
		\label{fig:seuils}
	\end{center}
\end{figure}

La figure \ref{fig:seuils} illustre l'étape qui consiste à définir des seuils ($S$) sur l'espace de la variable à régresser.

\subsection*{A-2 : Illustration de l'extension du vecteur $X$}

\begin{figure}[h]
	\begin{center}
		\includegraphics[width=0.75\textwidth]{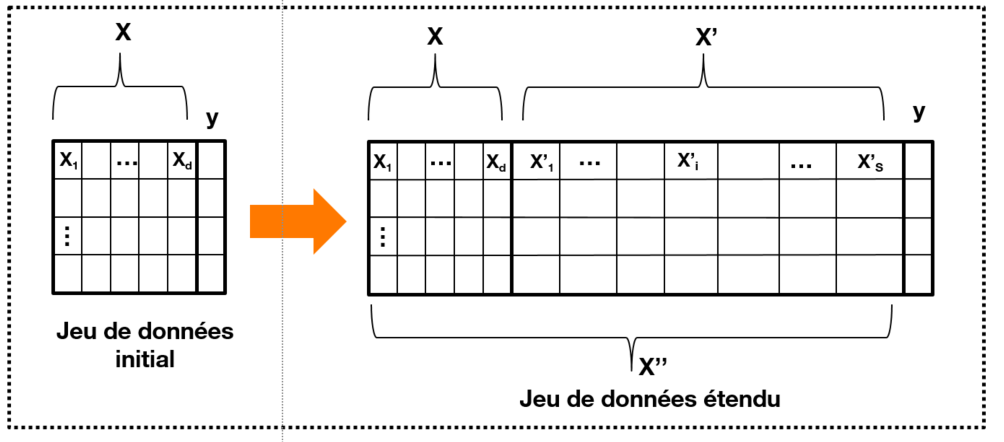}
		\caption{Extension de l'ensemble de données.}
		\label{fig:DatasetExtension}
	\end{center}
\end{figure}

La figure \ref{fig:DatasetExtension} illustre l'extension du vecteur $X$ (ayant $d$ composantes) au vecteur $X''$ (ayant $d+S$ composantes). 

\subsection*{A-3 : Schématisation du principe général de la méthode}

\begin{figure}[H]
	\begin{center}
		\includegraphics[width=0.75\textwidth]{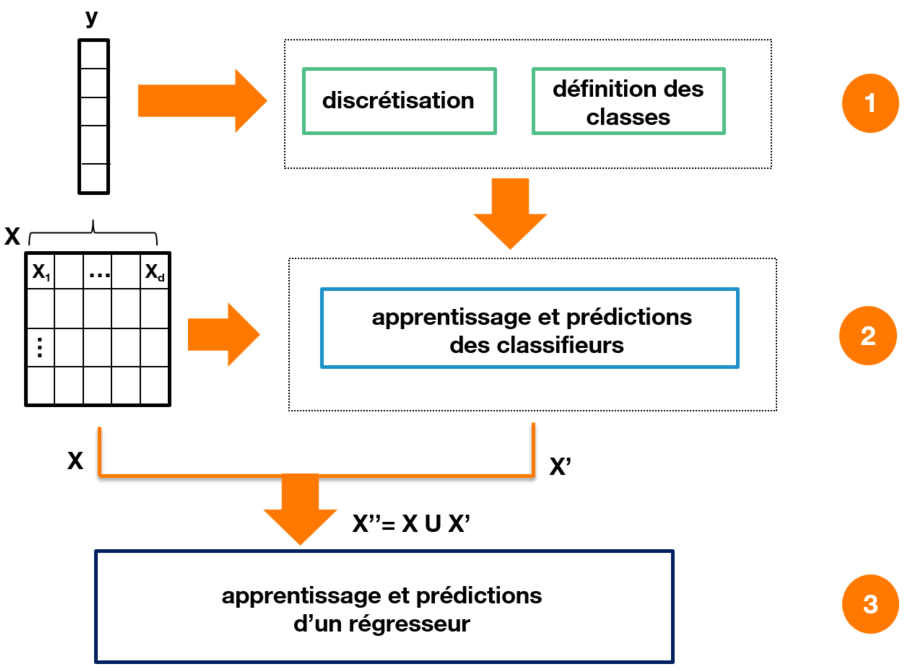}
		\caption{Schématisation du principe général de la méthode.}
		\label{fig:generalprinciple}
	\end{center}
\end{figure}

La figure \ref{fig:generalprinciple} résume le processus suivi par la méthode proposée.

\subsection*{A-4 : Détails suppleméntaires sur les variables extraites ($X'$)}

Pour la phase expérimentale qui a été mise en place, les variables extraites par la méthode proposée sont les probabilités conditionnelles prédites par chaque classifieur. Plus précisément, chaque classifieur a été entraîné à prédire si les données que l'on lui fournit sont inférieures ou supérieures au seuil auquel il est associé. Les probabilités conditionnelles de la classe 1 (c'est a dire si $y \geq y_i$) prédites par chaque classifieur composeront donc le vecteur $X'=\{X'_1, ..., X'_i, ..., X'_S\}$ (avec $S$ le nombre de seuils) qui sera ajouté à l'ensemble de données initial.

Pour pouvoir extraire cette probabilité conditionnelle, il faut donc un classifieur capable de générer ce genre de résultats. Dans le cas des forêts aléatoires qui ont été utilisées comme modèle de classification au sein de notre méthode pour les résultats présentés dans cet article, la probabilité conditionnelle correspond à la proportion d'arbres ayant voté pour une certaine classe. Avec 100 arbres dans les forêts qui ont été définies, la précision de la prédiction sera donc à 0.01 près.

\subsection*{A-5 : Détails sur les jeux de données et prétraitements}
\label{datasets}

\begin{table*}[!h]
    \centering
    \fontsize{7}{9}\selectfont
    \setlength{\tabcolsep}{2pt}
    \begin{tabular}{|l|l|l|l||l|l|l|l|}
        \hline
        Jeu de données       & \#individus    & \# variables & target & Jeu de données       & \#individus    & \# variables & target\\ \hline
        \hline
        3Droad               & 434,874        & 3       & altitude   & geo-lat              & 1,059          & 116     & latitude\\ \hline
        air-quality-CO       & 1,230          & 8       & PT08.S1(CO)  & greenhouse-net       & 955,167        & 15      & synthetic var  \\ \hline
        airfoil              & 1,503          & 5       & scaled sound  & KEGG-reaction        & 65,554         & 27      & edge count\\ \hline
        appliances-energy    & 19,735         & 26      & appliances    &   KEGG-relation        & 54,413         & 22      & clustering coef  \\ \hline
        beijing-pm25         & 41,758         & 14      & PM2.5 & online-news          & 39,644         & 59      & shares \\ \hline
        temp-forecast-bias   & 7,752          & 22      & Next\_Tmax &  video-transcode      & 68,784         & 26      & utime\\ \hline
        bike-hour            & 17,379         & 17      & count  &  pm25-chengdu-us      & 27,368         & 20      & PM\_US Post\\ \hline
        blog-feedback        & 60,021         & 18      & target  &  park-total-UPDRS     & 5,875          & 16      & total\_UPDRS\\ \hline
        buzz-twitter         & 583,250        & 70      & discussions &  physico-protein      & 45,730         & 9       & RMSD \\ \hline
        combined-cycle       & 9,568          & 4       & PE   & production-quality   & 29,184         & 17      & quality \\ \hline
        com-crime            & 1,994          & 122     & Violent crimes &  CT-slices            & 53,500         & 384     & reference  \\ \hline
        com-crime-unnorm     & 2,215          & 134     & Violent crimes  & gpu-kernel-perf      & 241,600        & 14      & Run1 (ms) \\ \hline
        compress-stren       & 1,030          & 8       & compressive strength & SML2010              & 4,137          & 26      & indoor temp \\ \hline
        cond-turbine         & 11,934         & 15      & gt turbine & seoul-bike-sharing   & 8,760          & 9      & Rented Bike Count  \\ \hline
        cuff-less            & 61,000         & 2       & APB  & uber-location-price  & 205,670        & 5       & fare amount\\ \hline
        electrical-grid-stab & 10,000         & 12      & stab  &  year-prediction      & 515,345        & 90      & Year \\ \hline
        facebook-comment     & 40,949         & 53      & target & & & & \\ \hline
	\end{tabular}
    \caption{Description des 33 jeux de données. La première colonne  désigne le nom du jeu (ou du sous-jeu) de données provenant de l'UCI ou de Kaggle.  La deuxième et la troisième colonne donnent respectivement le nombre d'individus  et le nombre de variables (explicatives) initial du jeu de données (après les étapes de prétraitements). Enfin la quatrième colonne indique le nom de la variable cible à régresser.}
	\label{table:DatasetsDescription}
\end{table*}

Pour réaliser l'analyse de la méthode proposée, nous avons sélectionné une large collection de jeux de données de régression provenant du UCI Repository \citep{Dua:2019} ou de Kaggle. 
Au total, 33 jeux de données ont été utilisés, parmi lesquels 23 sont constitués de plus de 10 000 individus. Les 10 jeux de données restants  allant de 1 030 à 9 568 individus (voir Table \ref{table:DatasetsDescription}). Certains de ces jeux de données comprenaient plusieurs sous-ensembles de données avec différentes variables cibles à régresser, et donc différents problèmes de régression. Il a été choisi de ne pas inclure  plusieurs problèmes de régression provenant des mêmes ensembles de données, afin de pas donner plus de poids à certains ensembles de données et créer un biais dans la comparaison des méthodes qui pourrait favoriser une méthode par rapport à une autre. Le lecteur pourra se référer à la colonne "nom de l'ensemble de données" de la Table \ref{table:DatasetsDescription} qui désigne le problème de régression utilisé pour chaque jeu de données.

\subsection*{A-6 : Détails supplémentaires sur les régresseurs utilisés}

Lors des expérimentations, cinq régresseurs utilisant des cadres différents ont été utilisés. Ils sont cités dans l'article. Il sont davantage décrits  ci-dessous, tout en étant largement connus par la communauté de l'apprentissage automatique.

{$\bullet$ Régression Linéaire (LR\footnote{Dans la suite les acronymes anglais seront utilisés.}) : }
La régression désigne le processus d'estimation d'une variable numérique continue à l'aide d'autres variables qui lui sont corrélées. Cela signifie que les modèles de régression prennent la forme de $y=f(X)$ où $y$ peut prendre un ensemble infini de valeurs. Dans cet article, nous avons choisis une régression linéaire multivariée dont le modèle, $f$, est déterminé par la méthode des moindres carrés. Nous avons utilisé la version 0.24.2 de Scikit-Learn. 
Ce modèle ne nécessite pas d'optimisation de ses paramètres.


{$\bullet$ Arbre de Régression (DT) : }
Les arbres de décision permettent de prédire une quantité réelle qui, dans le cas de régression, est une valeur numérique. Usuellement, les algorithmes pour construire les arbres de décision sont construits en divisant l'arbre du sommet vers les feuilles en choisissant à chaque étape une variable d'entrée qui réalise le meilleur partage de l'ensemble d'objets. 
Dans le cas des arbres de régression, on cherche à maximiser la variance inter-classes (avoir des sous-ensembles dont les valeurs de la variable-cible sont le plus dispersées possibles). Dans cet article, nous avons utilisé la version 0.24.2 de Scikit-Learn. Ce modèle  nécessite une optimisation de ses paramètres.

{$\bullet$ Forêt aléatoire (RF) : } Les forêts d'arbres décisionnels (ou forêts aléatoires de l'anglais Random Forest (RF) régresseur) ont été  proposées par Ho en 1995 
[Ho, T. K. (1995). Random decision forests. In Proceedings of 3rd International Conference
on Document Analysis and Recognition, Volume 1, pp. 278–282 vol.1.]
puis étendu par Leo Breiman et Adele Cutler [Breiman, L. (2001). Random forests. Machine Learning 45, 5–32.]
Cet algorithme combine les concepts de sous-espaces aléatoires et de bagging. L'algorithme des forêts d'arbres décisionnels effectue un apprentissage sur de multiples arbres de décision entraînés sur des sous-ensembles de données légèrement différents. Dans cet article, nous avons utilisé la version 0.24.2 de Scikit-Learn. Ce modèle  nécessite une optimisation de ses paramètres.

{$\bullet$ XGBoost (XGB): }
XGBoost [Chen, T. et C. Guestrin (2016). Xgboost : A scalable tree boosting system. In the 22nd ACM
SIGKDD International Conference, pp. 785–794.]
est une méthode de boosting. Elle combine séquentiellement des apprenants faibles qui  présenteraient individuellement de mauvaises performances pour améliorer la prédiction de l'algorithme complet. Un apprenant faible est un régresseur qui a une faible performance de régression. Dans cet algorithme de boosting, des poids élevés seront associés aux apprenants faibles ayant une bonne précision, et à l'inverse, des poids plus faibles aux apprenants faibles ayant une mauvaise précision. Dans la phase d'entraînement, des poids élevés sont associés aux données qui ont été mal "apprises" afin que l'apprenant faible suivant dans la séquence se concentre davantage sur ces données. Dans cet article, nous avons utilisé la version 1.4.1 de XGBoost. Ce modèle  nécessite une optimisation de ses paramètres.

{$\bullet$  Na\"ive Bayes Pondéré (SNB): }
Dans le contexte de la régression, un régresseur naïf de Bayes (NB) dont les variables sont pondérées par des poids (NBP) peut être obtenu à l'aide de deux étapes.
Dans un premier temps, pour chaque variable explicative, on crée une grille 2D pour estimer $P(X,y)$ (voir par exemple [Boullé, M. (2006). MODL : a Bayes optimal discretization method for continuous attributes. Machine Learning 65(1), 131–165.])
Ensuite, dans une deuxième étape, toutes les variables sont regroupées dans un algorithme Forward Backward 
pour estimer leur informativité dans le contexte d'un régresseur Naïf de  Bayes. A la fin de la deuxième étape, le modèle final (à déployer) est un Na\"ive Bayes (qui utilise la discrétisation 2D trouvée dans la première étape) où les variables sont pondérées (les poids sont trouvés dans la deuxième étape). Dans cet article nous avons utilisé le Na\"ive Bayes Pondéré produit par le logiciel Khiops  ({\url{www.khiops.com}}). Nous avons demandé et obtenu gratuitement une licence provisoire du logiciel  ainsi que sa version pyKhiops utilisable dans Scikit-Learn. Ce modèle ne nécessite pas d'optimisation de ses paramètres.

\end{document}